# Statistical language learning

**Eugene Charniak**
(Brown University)

*Reviewed by*
*David M. Magerman*
*Bolt Beranek and Newman Inc*

## 1. Introduction

The $64,000 question in computational linguistics these days is: "What should I read
to learn about statistical natural language processing?" I have been asked this question
over and over, and each time I have given basically the same reply: there is no text that
addresses this topic directly, and the best one can do is find a good probability-theory
textbook and a good information-theory textbook, and supplement those texts with an
assortment of conference papers and journal articles. Understanding the disappoint-
ment this answer provoked, I was delighted to hear that someone had finally written a
book directly addressing this topic. However, after reading Eugene Charniak's *Statistical
Language Learning*, I have very mixed feelings about the impact this book might have on
the ever-growing field of statistical NLP.

The book begins with a very brief description of the classic artificial intelligence
approach to NLP (chapter 1), including morphology, syntax, semantics, and pragmatics.
It presents a few definitions from probability theory and information theory (chapter
2), then proceeds to introduce hidden Markov models (chapters 3–4) and probabilistic
context-free grammars (chapters 5–6). The book concludes with a few chapters dis-
cussing advanced topics in statistical language learning, such as grammar induction
(chapter 7), syntactic disambiguation (chapter 8), word clustering (chapter 9), and word
sense disambiguation (chapter 10).

To its credit, the book serves as an interesting popular discussion of statistical mod-
eling in NLP. It is well-written and entertaining, and very accessible to the reader with
a limited mathematical background. It presents a good selection of statistical NLP top-
ics to introduce the reader to the field. And the descriptions of the forward-backward
algorithm for hidden Markov models and the inside-outside algorithm for probabilistic
context-free grammars are intuitive and easy to follow.

However, as a resource for someone interested in entering this area of research, this
book falls far short of its author's goals. These goals are clearly stated in the preface:

> This book aims to acquaint the reader with both the current state of
> the art in statistical language processing … and the mathematical back-
> ground necessary to understand the field. …Because the book is rela-
> tively self-contained, it could be used at the intermediate or advanced
> undergraduate level. However, because of its narrow focus, and because
> it skirts so close to the research edge in this area, it would probably be
> better as a beginning graduate-level course. (p. xvii)

The book is written at a level that is perhaps appropriate for an undergraduate student. But it is too brief (170 pages) for a book that claims to be a self-contained summary of the current state of the art in this field. In particular, far too little attention is paid to the mathematical fields of study upon which the statistical methods are based: probability theory and information theory. The book contains too few references to the literature (44 bibliography entries) to be useful as a springboard to other sources for a researcher. And it is neither self-contained nor an adequate presentation of the state of the art, even as it existed in 1993 when the book was written.

One of the mixed blessings of this book is the tone of the technical discussions. Charniak is trying to appeal to the widest possible audience in his book, and he is to be applauded for his effort. However, in avoiding mathematical formalism and concrete proofs in favor of intuitive explanations and simple examples, he withholds from the reader the critical building blocks of this field. While the technical material is generally accurate, the intuitive discussions are frequently off the mark. If the reader were offered the mathematical proofs, he or she could evaluate the discussions on their merits. However, without the technical details, the reader must place faith in the author's intuition.

My main concern about this book is that it will be taken as the definitive presentation of statistical NLP. It is the first book on the topic, and the only one currently available. Given the exposure it is likely to receive as a result, it might be worthwhile to explore some of the shortcomings of the book in more detail. In particular, I will address three main points: (1) the limited mathematical background provided in the book, (2) the gaps in the book's coverage of the literature, and (3) some significant oversimplifications made in the book.

## 2. "A fragment of probability theory"

One of the book's aims is to acquaint the reader with "the mathematical background necessary to understand the field." This mathematical background is no less than probability theory and information theory, and each of these topics demands at least a chapter, if not an entire textbook. In a book about statistical modeling, more discussion of probability theory is warranted than the three-page section entitled "A fragment of probability theory", which is all this book devotes to the topic. Also, a formal presentation of the concepts of information theory and their relationship to one another would be much more informative than the casual treatment they are given here. Readers may appreciate and enjoy the chatty tone of the examples given in the chapter, but it is necessary to understand these concepts at a formal mathematical level in order to comprehend the topics discussed later in the book, as well as other topics that are not mentioned but are relevant to the field.

In particular, the book omits a number of important definitions. Significantly, there is no formal definition of what constitutes a probability model! Nowhere in the text does it say that a probability function is a mapping onto the closed set of real numbers between 0 and 1. Nor does it state that, in the discrete case, the probabilities of all possibilities must sum to 1. This latter detail, however, is necessary to solve at least one of the exercises. The book also uses notation like $P(X = x, Y = y)$ without mentioning the term *joint probability* or defining the concept it represents. And, while information-theoretic concepts such as entropy and cross entropy are defined in some detail early in the book, the definitions of other important concepts, such as conditional entropy, relative entropy, and mutual information, are scattered throughout the text, without a clear description of the relationship among them. Many of the measures in information theory can be derived from combinations of other measures, and it is very instructive to see these derivations.

One of the consequences of the lack of mathematical background presented in the book is that it makes it difficult for students to identify careless mistakes in the literature, such as typographical errors in formulas and mathematical errors in derivations. For instance, in Chapter 2, the application of the source-channel model to the speech-recognition problem is introduced, where the probability of the transcription of an utterance, $w_{1,n}$, given the speech signal is estimated using Bayes' rule:

$$P(w_{1,n}|\textbf{speech signal}) = \frac{P(w_{1,n})P(\textbf{speech signal}|w_{1,n})}{P(\textbf{speech signal})}.$$

Later, in a discussion of an example of the speech-recognition algorithm, the text states that recognizers search for the sequence of words that maximizes the quantity:[1]

$$P(a_2, b_1, c_4)P(a_2, b_1, c_4|\textbf{speech signal}).$$

The correct formula is actually:

$$P(a_2, b_1, c_4)P(\textbf{speech signal}|a_2, b_1, c_4).$$

This is most certainly a typographical error, although it is repeated twice. But the reader, without confidence in his or her understanding of probability theory, is not likely to identify this as an error.

One might suggest that the author is of course assuming that the student will augment his or her reading with an appropriate probability-theory text and a good introduction to information theory. However, he fails to recommend a good resource for these fields of study, of which there are many (e.g., DeGroot 1986 or Bickel and Doksum 1977 for probability theory and Cover and Thomas 1991 or Ash 1965 for information theory), or to include any such texts in his bibliography. It seems as though he feels his presentation is sufficient to understand the material, or at least that further discussion of these topics is unnecessary.

This treatment of the mathematical foundations of statistical modeling is common in the statistical NLP literature, and it is motivated by a misinterpretation of the role statistical methods play in this field of research. Statistical methods are not simply well-understood tools that are used to learn about language; they are one of the central concerns of the research. Improvements in language modeling have not been achieved by viewing a trigram model as a black box and trying to work around its deficiencies. Over the last five years, language models have been improved by discovering precisely why $n$-gram models are estimating the probabilities of certain events poorly (generally because of sparse data) and finding alternative techniques that are more accurate. There are many different approaches to this problem, but all of them require a detailed understanding of the fundamentals of at least probability theory, if not information theory as well. A textbook that suggests that this understanding is unnecessary, or that it can be achieved by reading a few pages, is misleading at best.

### 3. Surveying the state of the art

Evaluating and presenting the state of the art in statistical NLP is more difficult than in more established fields because there are so few concentrated sources of material to point to and summarize. In the absence of books on the subject, the best places to look are journals, conference proceedings, and *published* workshop proceedings. Another good resource is recent doctoral theses, although these tend to be more verbose and overly technical, and they are frequently summarized later in journals.

---

1 The denominator drops out of the equation, since $P(\textbf{speech signal})$ remains constant when the word sequence is varied.

Conveniently, much of the foundational work in statistical NLP has been published in the proceedings of the Speech and Natural Language Workshop (later called the Human Language Technologies Workshop), sponsored by DARPA (later called ARPA) (DARPA 1989a, 1989b, 1990, 1991, 1992; ARPA, 1993, 1994). However, the book's bibliography fails to cite any papers from any of these workshops, many of which were important in the development of statistical NLP (e.g., Church *et al.* 1989, Gale and Church 1990, Brill *et al.* 1990, Chitrao and Grishman 1990, Magerman and Marcus 1991, Black *et al.* 1992a, 1992b, Brill 1992, and Lau 1993). Of the 44 bibliography entries, only two papers from *Computational Linguistics* are mentioned, omitting papers such as Brown *et al.* 1990, Seneff 1992 and Hindle and Root 1993 footnoteHindle and Root 1993 is an expanded, journal-length version of a conference paper by the same title, which is cited in the book. among others. And from the major computational linguistics conferences, whose participants have recently complained about the overwhelming number of papers on statistical methods, only eight papers are cited.

There are a number of papers cited from the working notes of two AAAI workshops: the 1992 AAAI Fall Symposium Series on Probabilistic Approaches to Natural Language and the 1992 AAAI Workshop on Statistically-Based NLP Techniques. These papers may be interesting and worthy of discussion, but they are of less value to the reader than papers from published proceedings and journals, since working notes are much more difficult to access for those who did not attend the workshops.

The omissions of important papers and published resources are a problem in the book; but they are symptomatic of a general lack of coverage of the mainstream literature throughout. Some of the important and foundational papers are discussed in the book, but many others are ignored, replaced by discussions of papers that are more on the periphery of the field. And many of the omitted papers represent the state of the art in the fields of language modeling, part-of-speech tagging, parsing, and general statistical language learning.

Some recently published work relevant to the study of these areas include: Lafferty, Sleator, and Temperly 1992, Lau, Rosenfeld, and Roukos 1993, and Della Pietra *et al.* 1994 for language modeling, Brill 1993 and Merialdo 1994 for part-of-speech tagging, Bod 1993 and Magerman 1994 for parsing, and Resnik 1993, Miller *et al.* 1994, and Yarowsky 1994 for various topics in statistical language learning.

### 4. Oversimplifications

In an introductory text, it is advisable to simplify some concepts for the reader to avoid confusion. Charniak employs this technique to great advantage. Occasionally, however, he crosses the line from simplification to oversimplification, leading the reader to draw inappropriate conclusions.

One example is the book's treatment of smoothing. Smoothing probability models in the face of sparse data is a hot topic in the field, and it is worthy of far more discussion than it is given in this book. One technique, deleted interpolation, is discussed at length.[2] But after this technique is introduced, explained, and discussed in Chapter 3, it is ignored

---

2 Actually, the book never uses the term *deleted interpolation* when describing the technique. This omission makes it difficult for the reader to investigate this technique further in the literature.

in future discussions. Consider the following passage:

> It is also useful to consider some not-so-good language models to see why they are less desirable. ….[Consider a model in which] the next part of speech is conditioned on not just the previous part of speech, but the previous word as well. In the abstract, including this dependence might or might not help the word model. In actuality, it is almost certainly a bad idea because of its effect on the sparse-data problem. (p. 48)

This is an excellent opportunity to suggest that smoothing this distribution with deleted interpolation would alleviate the sparse data problem, but no mention of this alternative is made.

Later, in discussing the prospects of estimating a 4-gram model for modeling prepositional phrase attachment,

$$P(\text{attachment} \mid \text{verb, noun, preposition, object of preposition}),$$

the text states plainly, "Unfortunately, this is still far from being something for which we can collect statistics," (p. 120). Again, no mention is made of the potential application of deleted interpolation or any other smoothing techniques. The only proposed solutions involve reducing the number of parameters in the model in ways that make the parameters estimable directly from frequency counts from a corpus. These reductions yield much weaker models and are unnecessary given current modeling techniques.

Another example of oversimplification occurs in the text's presentation of deleted interpolation. The definition of deleted interpolation applied to smoothing trigram word models is technically correct:

$$
\begin{aligned}
P(w_n|w_{n-1}w_{n-2}) \;=\; & \lambda_1 P_e(w_n) + \\
& \lambda_2 P_e(w_n|w_{n-1}) + \\
& \lambda_3 P_e(w_n|w_{n-1}w_{n-2}).
\end{aligned}
$$

Here, the $\lambda$ parameters are simply scalar values. However, in Brown *et al.* 1992 (also in Bahl, Jelinek, Mercer 1983) where deleted interpolation is defined, these parameters represent *functions* of the history, $\lambda(w_{n-1}w_{n-2})$ :

$$
\begin{aligned}
P(w_n|w_{n-1}w_{n-2}) \;=\; & \lambda_1(w_{n-1}w_{n-2})P_e(w_n) + \\
& \lambda_2(w_{n-1}w_{n-2})P_e(w_n|w_{n-1}) + \\
& \lambda_3(w_{n-1}w_{n-2})P_e(w_n|w_{n-1}w_{n-2}).
\end{aligned}
$$

This may seem like a minor point, but the effectiveness of deleted interpolation depends on this distinction, and the repercussions of this oversimplification are quite apparent when evaluating the performance of language models using each of these definitions.

To illustrate this, consider the example used in the text, trigram language modeling. A trigram language model trained on a large corpus using the scalar parameters would yield the result that *on average* the empirical estimates of the trigram model are more reliable than that of the bigram model, and that both are more reliable on average than the unigram model. But there are many bigram contexts, or *histories*, $(w_{n-1}w_{n-2})$ that are infrequent, and in these cases the trigram model is unreliable. Since the trigram model cannot be relied upon in all cases, the bigram and unigram model parameters ($\lambda_2$ and $\lambda_1$, respectively), will never be close to zero. For the sake of discussion, let's use the author's guesses of these parameters, since they are quite reasonable: $\lambda_3 = 0.6$, $\lambda_2 = 0.3$, and $\lambda_1 = 0.1$.

For the most frequent bigram histories, the empirical trigram model is the best predictor of the next word. For instance, the probability of the word that follows the bigram **in the** can be accurately estimated by an empirical trigram model. In contrast, the probability of the word following **volcanologic astrobiology** is better estimated using a unigram model. The formulation of the deleted interpolation parameters of Brown *et al.* (1992) is designed to allow high-frequency histories to depend on the trigram counts while deferring to the bigram and unigram counts for the low-frequency histories. So, $\lambda_3$(**in the**) will be very close to 1. Using the book's formulation, $P$(**the**|**in the**) is significantly overestimated because the model derives 10% of its estimate from the unigram model, even though the direct estimate from corpus frequencies is more accurate. Deleted interpolation as described by Brown *et al.* (1992) yields a far better language model than the simpler formulation used in the book, in terms of both entropy and performance.

This oversimplification of deleted interpolation occurs frequently in the literature. In fact, it probably occurs more frequently than can be determined, since details about smoothing algorithms are often omitted from conference papers. However, it is especially important that this concept be spelled out clearly in a textbook. Charniak does allude to the full definition of deleted interpolation in exercise 3.2, where he suggests the possibility of interpreting the $\lambda$s of a trigram model as functions of the word bigram history. However, this exercise is never answered or discussed in the text, and the topic is never brought up again. This point is far too critical to be relegated to an unsolved exercise. Glossing over this issue only serves to perpetuate the misuse of deleted interpolation in published research.

## 5. Summary

I cannot recommend this book without strong qualifications. It is the only book available that discusses this field at any length, and it is one of the few presentations of this material that is both substantive and accessible to the non-expert. However, it fails to accomplish even its own stated goals, much less satisfy the needs of the community at large.

For the casual reader interested in a snapshot of this field for the sake of personal knowledge, this book is quite adequate. It is not necessarily the snapshot I would have taken, but it certainly represents some of the work going on in the field in the past decade. One might place undergraduates in this category, in which case this book could be used in an elective NLP course for CS majors.

But many will attempt to use this book for other purposes. Some will teach graduate-level courses in statistical NLP using this book as a primary text. Others will offer this book to their students who are interested in exploring statistical NLP for their thesis work. And some researchers from different branches of NLP will read this book on their own, hoping to learn enough about statistical NLP to begin research in the field themselves.

To all of these potential readers, I offer the following advice: Start by reading the first few chapters of a probability-theory textbook, and the first few chapters of an information-theory textbook, in both cases focusing on the discussions of discrete (as opposed to continuous) models. You might read the middle chapters (3–6) of *Statistical Language Learning* for an initial introduction to topics such as hidden Markov models, the forward-backward algorithm and the inside-outside algorithm, in order to make the original sources more accessible. And, finally, study some of the articles and papers cited in this review for a better understanding of the applications of these techniques.

The overriding concern should be to learn (and teach) the mathematical underpinnings of the statistical techniques used in this field. The field of statistical NLP is very young, and the foundations are still being laid. Deep knowledge of the basic machinery is far more valuable than the details of the most recent unproven ideas.

*David Magerman* is a research scientist in the Speech and Natural Language Processing Department at Bolt Beranek and Newman Inc. He was a member of the IBM Speech Recognition group under Frederick Jelinek from 1991 to 1994, where he completed his doctoral thesis on statistical grammar acquisition. He also worked with Mitchell Marcus at the University of Pennsylvania from 1989 to 1991 on self-organized grammar learning and statistical parsing. Magerman's address is: Bolt Beranek and Newman Inc., Room 15/148, 70 Fawcett Street, Cambridge, MA 02138; e-mail: magerman@bbn.com.